\documentclass[journal]{IEEEtran}

\usepackage{mathptmx}
\usepackage{bbold}
\usepackage{graphicx}
\usepackage{times}
\usepackage{amssymb}
\usepackage{amsmath}
\usepackage{xcolor}
\usepackage{multirow}

\usepackage{algorithm}
\usepackage[noend]{algpseudocode}

\ifCLASSINFOpdf

\else

\fi

\begin{document}
\title{SPA-GAN: Spatial Attention GAN for Image-to-Image Translation}

\author{Hajar~Emami,
	Majid~Moradi Aliabadi,
	Ming~Dong,
    ~and Ratna~Babu Chinnam
	\thanks{H. Emami is with the Department of Computer Science, Wayne State University, Detroit, MI 48202. E-mail: hajar.emami.gohari@wayne.edu.}
	\thanks{M. Moradi Aliabadi is with the Department of Computer Science, Wayne State University, Detroit, MI 48202. E-mail: majid.moradi.aliabadi@wayne.edu.}
	\thanks{Corresponding author. M. Dong is with the Department of Computer Science, Wayne State University, Detroit, MI 48202. E-mail: mdong@wayne.edu.}
    \thanks{R. Chinnam is with the Department of Industrial and Systems Engineering, Wayne State University, Detroit, MI 48202. E-mail: Ratna.Chinnam@wayne.edu.}}

\maketitle

\begin{abstract}
Image-to-image translation is to learn a mapping between images from a source domain and images from a target domain. In this paper, we introduce the attention mechanism directly to the generative adversarial network (GAN) architecture and propose a novel spatial attention GAN model (SPA-GAN) for image-to-image translation tasks. SPA-GAN computes the attention in its discriminator and use it to help the generator focus more on the most discriminative regions between the source and target domains, leading to more realistic output images. We also find it helpful to introduce an additional feature map loss in SPA-GAN training to preserve domain specific features during translation. Compared with existing attention-guided GAN models, SPA-GAN is a lightweight model that does not need additional attention networks or supervision. Qualitative and quantitative comparison against state-of-the-art methods on benchmark datasets demonstrates the superior performance of SPA-GAN.
\end{abstract}


\IEEEpeerreviewmaketitle

\section{Introduction}
\label{intro}
\begin{figure*}[htpb]
\begin{center}
   \includegraphics[width=1\linewidth]{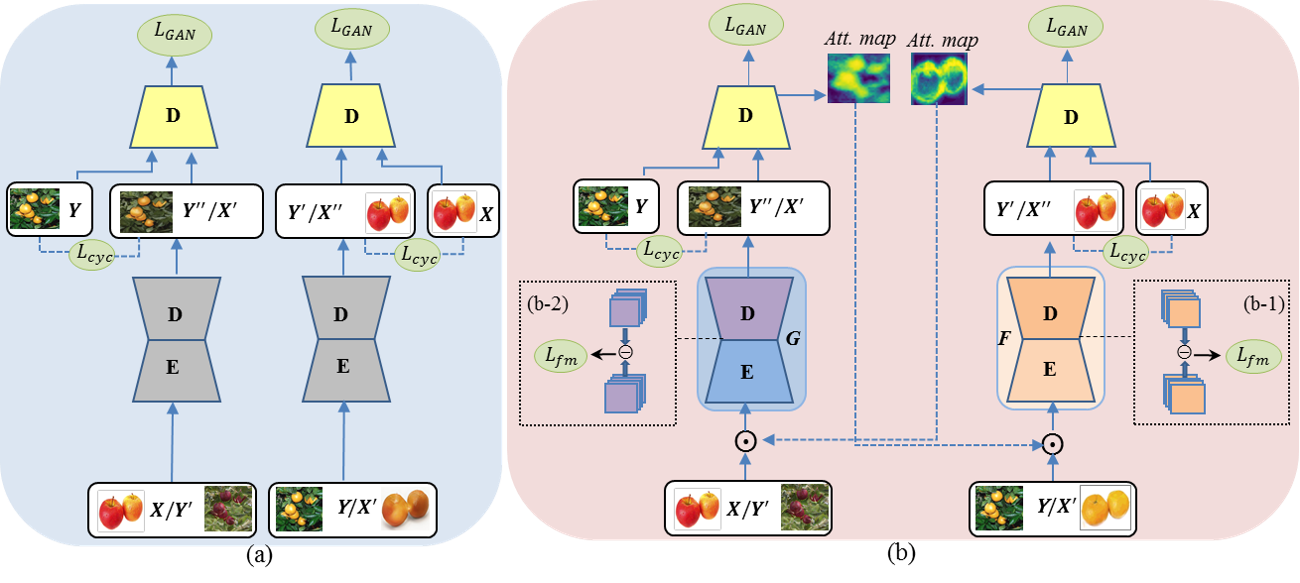}
\end{center}
      \vspace{-0.5cm}
   \caption{A comparison of CycleGAN (a) and SPA-GAN (b) architectures. In SPA-GAN, in addition to classifying the input images, the discriminator also generates spatial attention maps, which are fed to the generator and help it focus on the most discriminative object parts. In addition, the feature map loss is shown in the dashed blocks (b-1) and (b-2), which is the difference between the feature maps of the (attended) real and generated images computed in the first layer of the decoder. The feature map loss is used to preserve domain specific features in image translation.}
\label{fig:Fig1}
\end{figure*}

\IEEEPARstart{I}{mage-to-image} translation is to learn a mapping between images from a source domain and images from a target domain and has many applications including image colorization, generating semantic labels from images \cite{isola2017image}, image super resolution \cite{ledig2017photo, chen2019quality} and domain adaptation \cite{murez2018image}. Many image-to-image translation approaches require supervised learning settings in which pairs of corresponding source and target images are available. However, acquiring paired training data is expensive or sometimes impossible for diverse applications. Therefore, there are motivations towards approaches in unsupervised settings in which, source and target image sets are completely independent with no paired examples between the two domains. To this end, the need for paired training samples is removed by introducing the cycle consistency loss in unsupervised approaches \cite{zhu2017unpaired, yi2017dualgan} that force two mappings to be consistent with each other.
In general, an image-to-image translation method needs to detect areas of interest in the input image and learn how to translate the detected areas into the target domain. In an unsupervised setting with no paired images between the two domains, one must pay attention to the areas of the image that are subject to transfer. The task of locating areas of interest is more important in applications of image-to-image translation where the translation should be applied only to a particular type of object rather than the whole image. For example, for transferring an input ``orange'' image to the target domain ``apple'' (see the example in Fig.~\ref{fig:Fig1}), one needs to first locate the oranges in the input image and then transfer them to apples.

In \cite{zhu2017unpaired,yi2017dualgan}, a generative network is employed to detect areas of interest and translate between the two domains. Recent research shows that attention mechanism is helpful to improve the performance of generative adversarial networks (GANs) in image-to-image translation applications \cite{mejjati2018unsupervised, chen2018attention}. Specifically, the attention is introduced by decomposing the generative network into two separate networks: the attention network to predict regions of interest and the transformation network to transform the image from one domain to another. In \cite{mejjati2018unsupervised, chen2018attention}, additional attention networks are added to the CycleGAN framework to keep the background of the input image unchanged while translating the foreground. For example, Chen et al. \cite{chen2018attention} used the segmentation annotations of input images as extra supervision to train an attention network. Then, the attention maps are applied to the output of the transformation network so that the background of input image is used as the output background, leading to the improvement of the overall image translation quality.

In this paper, we introduce the attention mechanism directly to the GAN architecture and propose a novel spatial attention GAN model (SPA-GAN) for image-to-image translation. SPA-GAN computes the attention in its discriminator and use it to help the generator focus more on the most discriminative regions between the source and target domains. Specifically, the attention from the discriminator is defined as the spatial maps \cite{zagoruyko2016paying} showing the areas that the discriminator focuses on for classifying an input image as real or fake. The extracted spatial attention maps are fed back to the generator so that higher weights are given to the discriminative areas when computing the generator loss, leading to more realistic output images. In unsupervised setting, we also find it helpful to introduce an additional feature map loss to preserve domain specific features during translation. That is, in SPA-GAN's generative network, we constrain the feature maps obtained in the first layer of the decoder \cite{liu2016coupled} to be matched with the identified regions of interest from both real and generated images so that the generated images are more realistic. The major contribution of our work is summarized as follows:
\begin{itemize}
\item Different from \cite{mejjati2018unsupervised, chen2018attention} where attention is employed to separate foreground and background, we use attention in SPA-GAN as a mechanism of transferring knowledge from the discriminator back to the generator. The discriminator helps the generator explicitly attend to the discriminative regions between two domains, leading to more realistic output images.

\item Based on the proposed attention mechanism, we used a modified cycle consistency loss during SPA-GAN training and also introduced a novel generator feature map loss to preserve domain specific features during image translation.

\item Earlier approaches on attention-guided image-to-image translation \cite{mejjati2018unsupervised, chen2018attention} require loading generators, discriminators and additional attention networks into the GPU memory all at once, which may cause computational and memory limitations. In comparison, SPA-GAN is a lightweight model that does not need additional attention networks or supervision (e.g. segmentation labels) during training.

\item SPA-GAN demonstrates the effectiveness of directly incorporating the attention mechanism into GAN models. Through extensive experiments, we show that, both qualitatively and quantitatively, SPA-GAN significantly outperforms other state-of-the-art image-to-image translation methods on various benchmark datasets.

\end{itemize}
The remainder of this paper is organized as follows: Section II contains a brief review of the literature surrounding
image-to-image translation and attention learning. In Section III, we introduce our SPA-GAN model in detail. In Section IV, we present
our image-to-image translation results on benchmark datasets. Finally, we conclude in Section V.
\hfill

\hfill
\section{Related Work}
\subsection{Image-to-Image Translation}
Recently, GAN-based methods have been widely used in image-to-image translation and produced appealing results. In pix2pix \cite{isola2017image}, conditional GAN (cGAN) was used to learn a mapping from an input image to an output image; cGAN learns a conditional generative model using paired images from source and target domains. CycleGAN was proposed by Zhu et al. \cite{zhu2017unpaired} for image-to-image translation tasks in the absence of paired examples. It learns a mapping from a source domain $X$ to a target domain $Y$ (and vice versa) by introducing two cycle consistency losses. Similarly, DiscoGAN \cite{kim2017learning} and DualGAN \cite{yi2017dualgan} use an unsupervised learning approach for image-to-image translation based on unpaired data, but with different loss functions. HarmonicGAN proposed by Zhang et al. \cite{zhang2019harmonic} for unpaired image-to-image translation introduces spatial smoothing to enforce consistent mappings during translation.

More recently, Liu et al. \cite{liu2017unsupervised} proposed unsupervised image-to-image translation network (UNIT) based on Coupled GANs \cite{liu2016coupled} and the shared-latent space assumption that assumes a pair of corresponding images from different domains can be mapped to the same latent representation. Furthermore, some image-to-image translation methods assume that the latent space of images can be decomposed into a content space and a style space, which enables the generation of multi-modal outputs. To this end, Huang et al. \cite{huang2018multimodal} proposed multimodal unsupervised image-to-image translation framework (MUNIT) with two latent representations for style and content, respectively. To translate an image to another domain, its content code is combined with different style representations sampled from the target domain. Similarly, Lee et al.  \cite{lee2018diverse} introduced diverse image-to-image translation (DRIT) based on the disentangled representation on unpaired data that decomposes the latent space into two: a domain-invariant content space capturing shared information and a domain-specific attribute space to produce diverse outputs given the same content. Zhou et al. \cite{zhou2019branchgan} proposed BranchGAN to transfer an image of one domain to the other by exploiting the shared distribution of the two domains using the same encoder. InstaGAN \cite{mo2018instagan} utilizes the object segmentation masks as extra supervision to perform multi-instance domain-to-domain image translation. It preserves the background by introducing the context preserving loss. However, InstaGAN requires the semantic segmentation labels (i.e., pixel-wise annotation) for training and thus has limitation for applications where such information is not available.

\subsection{Attention Learning in Deep Networks}
Inspired from human attention mechanism \cite{rensink2000dynamic}, attention-based models have gained popularity in a variety of computer vision and machine learning tasks including neural machine translation \cite{bahdanau2014neural}, image classification \cite{mnih2014recurrent, wang2017residual}, image segmentation \cite{chen2016attention}, image and video captioning \cite{xu2015show, yao2015describing} and visual question answering \cite{yang2016stacked}. Attention improves the performance of all these tasks by encouraging the model to focus on the most relevant parts of the input. Zhou et al. \cite{zhou2016learning} produce attention maps for each class by removing top average-pooling layer and improving object localization accuracy. Zagoruyko et al. \cite{zagoruyko2016paying} improve the performance of a student convolutional neural network (CNN) by transferring the attention from a teacher CNN. Their scheme determines the attention map of a CNN based on the assumption that the absolute value of a hidden neuron activation is relative to the importance of that neuron in the task of classifying a given input. Minh et al. \cite{mnih2014recurrent} propose a visual attention model that is capable of extracting information from an image or video by adaptively selecting a sequence of regions or locations and only processing the selected regions at a high resolution. Kuen et al. \cite{kuen2016recurrent} propose a recurrent attentional convolutional-deconvolution network for saliency detection. This supervised model uses an iterative approach to attend to selected image sub-regions for saliency refinement in a progressive way. Wang et. al. \cite{wang2017residual} propose a residual attention network for image classification with a trunk-and-mask attention mechanism.

Recent studies show that incorporation of attention learning in GAN models leads to more realistic images in both image generation and image-to-image translation tasks. For example, Zhang et al. \cite{zhang2018self} propose self-attention GAN that uses a self-attention mechanism for image generation. In \cite{yang2017lr},  the LR-GAN model learns to generate image background and foregrounds separately and recursively, and the idea was later adapted to image-to-image translations. Specifically, Chen et al. \cite{chen2018attention} and Mejjati et al. \cite{mejjati2018unsupervised} add an attention network to each generator to locate the object of interest in image-to-image translation tasks. Since, the background is excluded from the translation, the quality of translated images in the background regions are improved in these approaches. However, improving the quality of translated objects and foreground is not the focus of these two approaches.
\section{SPA-GAN}
The goal of image-to-image translation is to learn a mapping $G$ from a source domain $X$ : $\big\{{x_i}\big\}_{i=1}^{N_X}$ to a target domain $Y$ : $\big\{{y_j}\big\}_{j=1}^{N_Y}$, where $N_X$ and $N_Y$ are the number of samples in domains $X$ and $Y$, respectively. In an unpaired setting, two inverse mappings are learned simultaneously through the cycle consistency loss \cite{zhu2017unpaired, yi2017dualgan}.

Incorporating the attention mechanism into image-to-image translations can help the generative network to attend to the regions of interest and produce more realistic images. The proposed SPA-GAN model achieves this by explicitly transferring the knowledge from the discriminator to the generator to force it focus on the discriminative areas of the source and the target domains. Fig.~\ref{fig:Fig1} shows the main components of SPA-GAN and compares it to the CycleGAN model with no feedback attention. Both frameworks learn two inverse mappings through one generator and one discriminator in each domain. However, in SPA-GAN the discriminator generates the attention maps in addition to classifying its input as real or fake. These attention maps are looped back to the input of the generator. While CycleGAN is trained using the adversarial and cycle consistency losses, SPA-GAN integrates the adversarial, modified cycle consistency and feature map losses to generate more realistic outputs.


\subsection{Spatial Attention Map from Discriminator}
In GAN, the discriminator classifies the input to either fake or real. In SPA-GAN, we deploy the discriminator network to highlight the most discriminative regions between real and fake images in addition to the classification. These discriminative regions illustrate the areas where the discriminator focuses on in order to correctly classify the input image.

Formally, given an input image $x$, the spatial attention map $A_{D_X}(x)$, whose size is the same as the input image $x$, is obtained by feeding $x$ to the discriminator. Following \cite{zagoruyko2016paying}, we define $A_{D_X}(x)$ as the sum of the absolute values of activation maps in each spatial location in a layer across the channel dimension:
\begin{align}
A_{D}&= \sum\limits_{i=1}^C |F_i|
\end{align}
where $F_i$ is $i$-th feature plane of a discriminator layer for the specific input and $C$ is the number of channels. ${A_{D}}$ directly indicates the importance of the hidden units at each spatial location in classifying the input image as a fake or real.

The attention maps of different layers in a classifier network focus on different features. For instance, when classifying apples or faces, the middle layer attention maps have higher activations on regions such as the top of an apple or eyes and lips of the face, while the attention maps of the later layers typically focus on full objects. Thus, in SPA-GAN we select the mid-level attention maps from the second to last layer in $D_X$, usually correlated to discriminative object parts \cite{zagoruyko2016paying}, and feed them back to the generator.

The detailed architecture of SPA-GAN is shown in panel (b) of Fig.~\ref{fig:Fig1}. First, an input image $x$ is fed to the discriminator $D$, to get the spatial attention map $A_{D_X}(x)$, the most discriminative regions in $x$. Then, the spatial attention map is normalized by dividing each value by the maximum value observed in the map and upsampeled to match the input image size. Next, we apply the spatial attention map to the input image $x$ using an element-wise product and feed it to the generator $G$ to help it focus on the most discriminative parts when generating $x^\prime$:
\begin{align}
x^\prime = G(x_a) = G(A_{D_X}(x) \odot x)
\end{align}
where $x_a$ is the attended input sample.

\subsection{Feature Map Loss}
Unsupervised image synthesizing requires two pairs of generator and discriminator as the mapping is done in both directions(see panel (b) of Fig.~\ref{fig:Fig1}). We make use of this architecture and also introduce an additional feature map loss term that encourages the generators to obtain domain specific features. Ideally, in the generator pair, both real and generated objects should share the same high-level abstraction in decoding. So, our feature map loss penalizes the differences between the feature maps in the first layer of decoders, which is responsible for decoding the high-level semantics \cite{liu2016coupled} in the real and generated images, respectively.

Specifically, the generator feature map loss between the attended sample $x_a$ in the source domain $X$ and the attended generated sample $y^{\prime}_a$ in the inverse mapping $Y^{\prime}$ is computed as follows (see dashed box (b-1) and (b-2) in Fig.~\ref{fig:Fig1}):
\begin{align}
{\cal{L}}_{fm}(G)&=  \frac{1}{C}\ \sum\limits_{i=1}^C(||G^i(x_a)-G^i(y^{\prime}_a)||_1)
\end{align}
where $G^i$ is the $i$-th feature map and C is the number of feature maps in the given layer of the generator $G$. The feature map loss computed in the first layer of decoders is added to the overall loss function of the generator $F$ to preserve domain specific features. The feature map loss associated with the inverse mapping $F: Y \rightarrow X$ can be defined similarly, and the total feature map loss is given as:
\begin{align}
{\cal{L}}_{fm}(G,F)&=  \frac{1}{C}\ \sum\limits_{i=1}^C(||G^i(x_a)-G^i(y^{\prime}_a)||_1)\nonumber\\
                               &+\frac{1}{C} \sum\limits_{i=1}^C (||F^i(y_a)-F^i(x^{\prime}_a)||_1)
\end{align}
where $||\cdot||_1$ is the $L_1$ norm. $L_1$ norm is widely used in image-to-image translation tasks. For instance, in \cite{isola2017image, zhu2017unpaired}, it was adopted because it leads to slightly less blurry results. $L_1$ was also used in \cite{wang2018high} to compute the discriminator feature matching loss and subsequently improve the image translation performance. Thus, we chose to use $L_1$ norm to compute our feature map loss. As shown in our experimental results in Section IV, the feature map loss helps generate more realistic objects by explicitly forcing the generators to maintain domain specific features.

\subsection{Loss Function}
The adversarial loss of GAN for the mapping $G:{X \rightarrow Y}$ and its discriminator $D_Y$ is expressed as:
\begin{align}
{\cal{L}}_{GAN}^x(G,D_Y, X, Y)&= \mathbb{E}_{y{\sim}Pdata(y)}[log(D_Y(y))]\nonumber\\
                              &+ \mathbb{E}_{x{\sim}Pdata(x)}[log(1-D_Y(G(x))]
\end{align}
and the inverse mapping $F:{Y \rightarrow X}$ has a similar adversarial loss:
\begin{align}
{\cal{L}}_{GAN}^y(F,D_X, Y, X)&=\mathbb{E}_{x{\sim}Pdata(x)}[log(D_X(x))]\nonumber\\
                                                &+\mathbb{E}_{y{\sim}Pdata(y)}[log(1-D_X(F(y))]
\end{align}
where the mapping functions $G$ and $F$ aim to minimize the loss against the adversary discriminators $D_Y$ and $D_X$ that try to maximize the loss.

A network with enough capacity might map a set of input images to any random permutation of images in the target domain, and thus the adversarial losses alone cannot guarantee a desired output $y$ from the input image $x$ with the learned mapping. To overcome this, Cycle consistency loss is proposed in CycleGAN \cite{zhu2017unpaired} to measure the discrepancy between the input image $x$ and the image $F(G(x))$ generated by the inverse mapping that translates the input image back to the original domain space. Similar to CycleGAN, we take advantage of cycle consistency loss to achieve one-to-one correspondence mapping. Since we apply the attention map extracted from the discriminator to the generator's input, we modify the cycle consistency loss as:
\begin{align}
{\cal{L}}_{cyc}(G,F)&=\mathbb{E}_{x{\sim}Pdata(x)}[||F(G(x_a))-x_a||_1]\nonumber\\
                    &+\mathbb{E}_{y{\sim}Pdata(y)}[||G(F(y_a))-y_a||_1]
\end{align}
where $x_a$ and $y_a$ are the attended input samples. The modified cycle consistency loss helps the generators to focus on the most discriminative regions in image-to-image translations. In \cite{mejjati2018unsupervised, chen2018attention}, the attended regions are the same for both mappings, and cycle consistency loss enforces the attended regions to conserve content (e.g., pose) of the object, which prevents the network from geometric and shape changes. Different from \cite{mejjati2018unsupervised, chen2018attention}, our framework allows different attention maps in the forward and inverse mappings.

Finally, by combining the adversarial loss, modified cycle consistency loss and the generator feature map loss, the full objective function of SPA-GAN is expressed as:
\begin{align}
{\cal{L}}(G,F,D_X,D_Y)&={\cal{L}}_{GAN}(G,D_Y, X, Y)\nonumber\\
                                 &+{\cal{L}}_{GAN}(F,D_X, Y, X)+\lambda_{cyc}{\cal{L}}_{cyc}(G,F)\nonumber\\
                &+\lambda_{fm}{\cal{L}}_{fm}(G,F)
\label{eq:8}
\end{align}
where $\lambda_{cyc}$ and $\lambda_{fm}$ control the importance of different terms, and we aim to solve the following min-max problem:
\begin{align}
G^*,F^*,D^*_X,D^*_Y &=\arg \, \smash{\displaystyle\min_{G,F}} \; \smash{\displaystyle\max_{D_X,D_Y}} {\cal{L}}(G,F,D_X,D_Y)
\end{align}

\section{Experiments}
In this section, we first perform ablation study of our model and analyze the effect of each component in SPA-GAN. Then, we compare SPA-GAN with current state-of-the-art methods on benchmark datasets qualitatively, quantitatively, and with user studies.

\subsection{Datasets and Experimental Setups}
We evaluate SPA-GAN on the Horse $\leftrightarrow$ Zebra, Apple $\leftrightarrow$ Orange datasets provided in \cite{zhu2017unpaired} and the Lion $\leftrightarrow$ Tiger dataset downloaded from ImageNet \cite{deng2009imagenet}, which consists $2,086$ images for tigers and $1,795$ images for lions. These are challenging image-to-image translation datasets including objects at different scales. The goal is to translate one particular type of object (e.g., orange) into another type of object (e.g., apple). We also evaluate SPA-GAN on image-to-image translation tasks that require to translate the whole image, e.g., Winter $\leftrightarrow$ Summer \cite{zhu2017unpaired}, gender conversion in the Facescrub \cite{ng2014data} dataset and the GTA \cite{richter2016playing} $\leftrightarrow$ Cityscapes \cite{cordts2016cityscapes} dataset.

For all experiments, we use the Adam solver \cite{kingma2014adam} and a batch size of 1. The networks were trained with an initial learning rate of 0.0002. We adopt the same architecture used in \cite{zhu2017unpaired} for our generative networks and discriminators. We use a least-squares loss \cite{mao2017least} which has been shown to lead to more stable training and help to generate higher quality and sharper images. Following \cite{zhu2017unpaired,mejjati2018unsupervised,chen2018attention}, we set $\lambda_{cyc}$ =10 in Eq.~\ref{eq:8}. We also tried $\lambda_{fm}$=1, 3 and 5 and empirically set it at 1, which gives the best performance based on KID. Different from \cite{mejjati2018unsupervised,chen2018attention} that add additional attention networks to the CycleGAN framework, SPA-GAN does not include any additional attention network or supervision, and its training time is similar to CycleGAN.

\subsection{Evaluation Metrics}
The following state-of-the-art image-to-image translation methods are used in our empirical evaluation and comparison.

\noindent \textbf{CycleGAN.} CycleGAN adopts GAN with cycle consistency loss for unpaired image-to-image translation task \cite{zhu2017unpaired}.\\
\textbf{DualGAN.} An unsupervised dual learning framework for image to image translation on unlabeled images from two domains that uses Wasserstein GAN loss rather than the sigmoid cross-entropy loss \cite{yi2017dualgan}. \\
\textbf{UNIT.} An unsupervised image-to-image translation framework based on the shared-latent space assumption and cycle loss \cite{liu2017unsupervised}. \\
\textbf{MUNIT.} A multimodal unsupervised image-to-image translation framework that assumes two latent representations for style and content. To translate an image to another domain, its content code is combined with different style representations sampled from the target domain  \cite{huang2018multimodal}. \\
\textbf{DRIT.} A diverse image-to-image translation approach based on the disentangled representation on unpaired data that decomposes the latent space into two: a domain-invariant content space capturing shared information and a domain-specific attribute space to produce diverse outputs given the same content. The number of output style is set to 1 in our experiments for both MUNIT and DRIT \cite{lee2018diverse}. \\
\textbf{Attention-GAN.} An unsupervised image-to-image translation method that decomposes the generative network into two separate networks: the attention network to predict regions of interest and the transformation network to transform the image from one domain to another \cite{chen2018attention}.\\
\textbf{AGGAN.} A similar unsupervised image-to-image translation method with added attention networks \cite{mejjati2018unsupervised}. It was reported that AGGAN outperformed the Attention-GAN \cite{chen2018attention}.

Two metrics, Kernel Inception Distance (KID) and classification accuracy, are used for quantitative comparison between SPA-GAN and the state-of-the-arts. KID \cite{binkowski2018demystifying} is defined as the squared Maximum Mean Discrepancy (MMD) between Inception representations of real and generated images. It has been recently used for performance evaluation of image-to-image translation and image generation models \cite{mejjati2018unsupervised, binkowski2018demystifying}. KID is an improved measure that has an unbiased estimator with no assumption about the form of activations distribution, which makes it a more reliable metric compared to the Fréchet Inception Distance (FID) \cite{heusel2017gans}, even for a small number of test samples. Smaller KID value indicates higher visual similarities between the generated images and the real images. Classification accuracy on the generated images is also widely used as a quantitative evaluation metric in image generation literature \cite{isola2017image, wang2016generative, zhang2016colorful}. In our experiment, we fine-tuned the inception network \cite{szegedy2016rethinking} pretrained on ImageNet \cite{deng2009imagenet} for each translation and report the top-1 classification performance on the images generated by each method. We also conducted a human perceptual study on different translation tasks to further evaluate our model.

\subsection{Ablation Study}
We first performed model ablation on the Apple $\rightarrow$ Orange dataset to evaluate the impact of each component of SPA-GAN. In Table~\ref{table:Table1}, we report both KID and classification accuracy for different configurations of our model. First, we removed the attention transfer from the discriminator to the generator (as a consequence, we also used the regular cycle consistency loss). The generator feature map loss is also removed because it is calculated only on the objects detected by the spatial attention map. In this case, our model is reduced to the CycleGAN architecture (CycleGAN). The KID and classification accuracy we obtained is consistent with the reported ones in the literature.

Next, we removed the attention transfer from the discriminator to the generator but kept feature map loss in the first layer of decoder (SPA-GAN-wo-$A_D$). Our results show that this leads to slightly better results (KID and classification accuracy) than CycleGAN, but much worse than our models with attention. This ablation study showed that feature map loss works better when spatial attention is employed as its computation can focus on the attended discriminative regions. Further, we fed the spatial attention from the discriminator to the generator in CycleGAN but without the generator feature map loss (SPA-GAN-wo-${\cal{L}}_{fm}$). Our results show that this leads to a higher KID and lower classification accuracy when compared with the full version of SPA-GAN. Clearly, by enforcing the similarity between the discriminative regions of the attended real image and the attended generated image, the feature map loss computed in the abstract level can help us achieve a more realistic output.
\begin{table}
\centering
\caption{Kernel Inception Distance $\times$ 100 $\pm$ std. $\times$ 100 (lower is better) computed using only the target domain and classification accuracy (higher is better) for ablations of our proposed approach on the Apple $\rightarrow$ Orange dataset.}.
\label{table:Table1}
\begin{tabular}{c|ccc}
\hline
   & KID & accuracy \\
\hline
CycleGAN &  11.02 $\pm$ 0.60 & 71.80\\
SPA-GAN-wo-$A_D$ & 10.44  $\pm$ 0.36 & 73.30\\
SPA-GAN-wo-${\cal{L}}_{fm}$ &  4.81 $\pm$ 0.23 & 85.71\\
SPA-GAN-${\cal{L}}_{fm}$-$E^1$ & 5.97 $\pm$ 0.52 & 78.90 \\
SPA-GAN-${\cal{L}}_{fm}$-$D^4$ & 5.49 $\pm$ 0.38 & 82.70  \\
SPA-GAN-$A_{max}$ & 5.66 $\pm$ 0.48 &  84.59 \\
SPA-GAN-${\cal{L}}_{fm}$-$D^1$ &  \textbf{3.77 $\pm$ 0.32} & \textbf{87.21}\\
\hline
\end{tabular}
\end{table}

\begin{figure}[htbp]
\centering
   \includegraphics[width=0.8\linewidth]{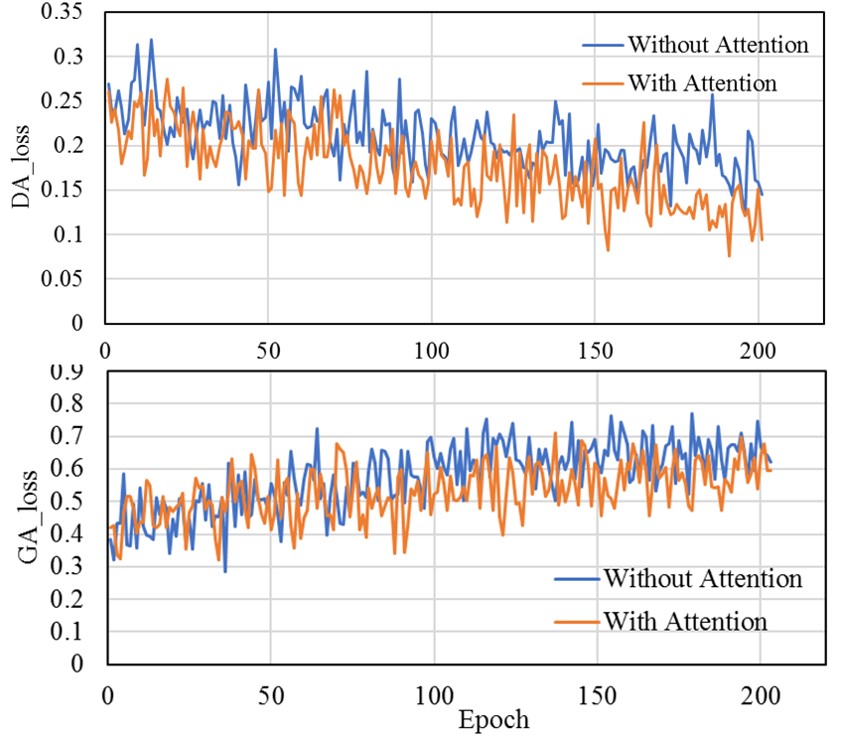}\\
      \vspace{-0.4cm}
   \caption{Comparison of training losses on the apple $\rightarrow$ orange task.}
\label{fig:Fig7}
\end{figure}

In our model ablation, we also compared different configurations of our model that calculate feature map loss on different layers of the generator, including the first layer of the encoder (SPA-GAN-${\cal{L}}_{fm}$-$E^1$), the first layer of the decoder (SPA-GAN-${\cal{L}}_{fm}$-$D^1$) and the fourth layer of the decoder (SPA-GAN-${\cal{L}}_{fm}$-$D^4$). Our results show that calculating feature map loss on the later layers (i.e., fourth layer) of the decoder (responsible for decoding low-level details) lead to a higher KID and lower classification accuracy than calculating it on the first layer of the decoder (responsible for decoding high-level semantics that are generic, abstract and discriminative \cite{liu2016coupled}). We also calculated feature map loss on the first layer of the encoder (SPA-GAN-${\cal{L}}_{fm}$-$E^1$). As shown in Table I, SPA-GAN-${\cal{L}}_{fm}$-$E^1$ got the higher KID and lower classification accuracy when compared with SPA-GAN-${\cal{L}}_{fm}$-$D^1$ and SPA-GAN-${\cal{L}}_{fm}$-$D^4$. This suggests that applying feature map loss on the embedding in the encoder does not improve image-to-image translation performance, possibly because embedding in the encoder is more related to the input images than the generated images. Overall, SPA-GAN-${\cal{L}}_{fm}$-$D^1$ achieved the best results in our model ablations, and for simplicity we denote it as SPA-GAN in the rest of our experiments.

As pointed out in \cite{zagoruyko2016paying}, attention can also be computed as the maximum of the absolute values in the activation maps. Thus, we also compared maximum-based attention (SPA-GAN-$A_{max}$) with sum-based attention adopted in SPA-GAN. Higher KID and lower classification accuracy of SPA-GAN-$A_{max}$ reported in Table~\ref{table:Table1} is consistent with the results in \cite{zagoruyko2016paying}. In the following experiments, we use the SPA-GAN with sum-based attention in our evaluation and compare it with exiting methods.

Finally, in Fig.~\ref{fig:Fig7}, we provide the training curves of both the generator and the discriminator of SPA-GAN and the translation model without attention mechanism (i.e., CycleGAN) for the apple $\rightarrow$ orange task. The training curves reveal that
SPA-GAN has lower generator and discriminator losses during training and also better convergence when compared with CycleGAN (i.e., mild oscillation in loss by SPA-GAN vs. more chaotic loss distribution by CycleGAN).

\subsection{Qualitative Results}
\begin{figure*}[htbp]
\begin{center}
   \includegraphics[width=0.8\linewidth]{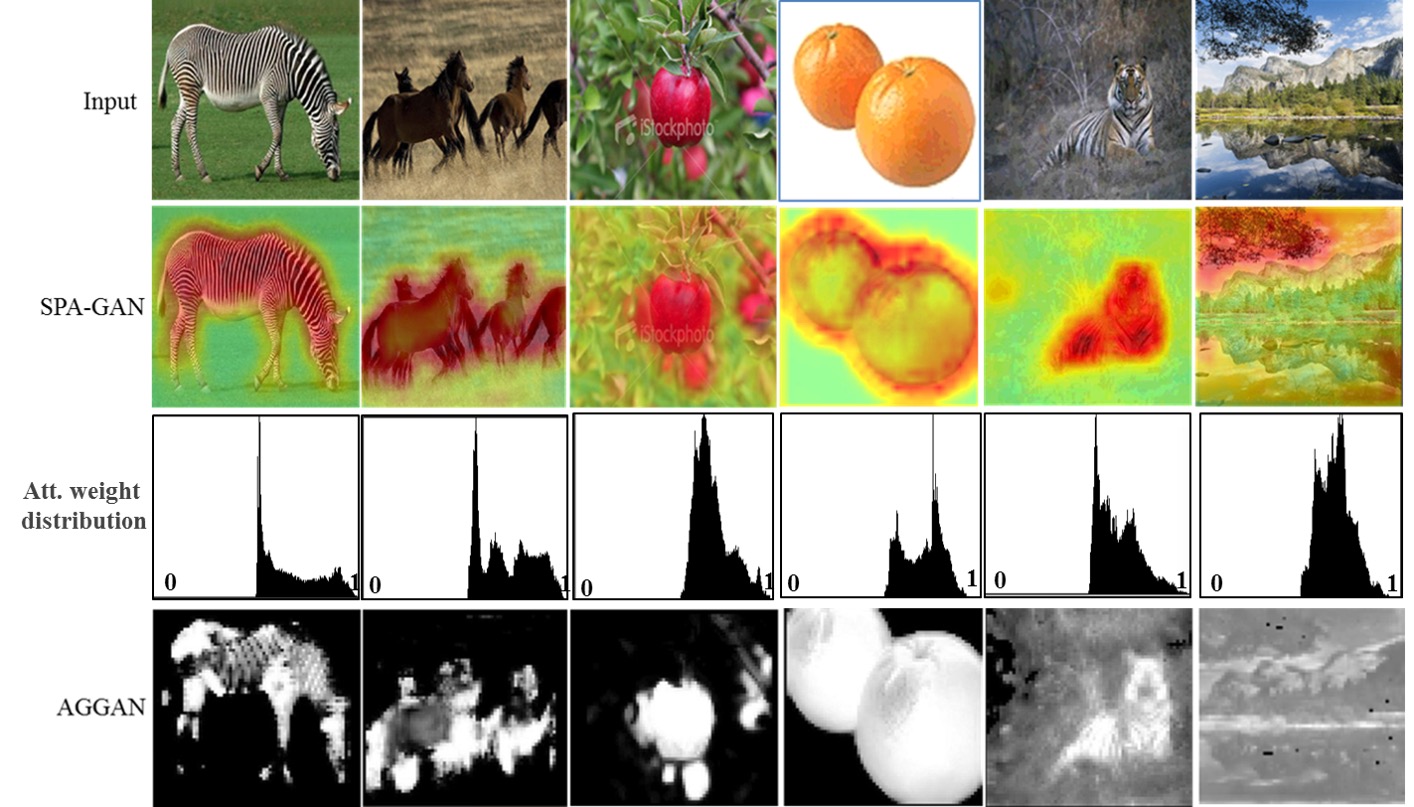}
         \setlength{\belowcaptionskip}{-20pt}
               \vspace{-0.3cm}
   \caption{Comparison between the attention maps generated by the attention network in AGGAN \cite{mejjati2018unsupervised} (the fourth row) and the attention maps computed in SPA-GAN (the second row) on different image samples. SPA-GAN attention maps have higher activation values in the most discriminative regions between the source and target domains. Note, for example, in column one AGGAN generates a disconnected attention map for zebra while SPA-GAN attends on all the zebra patterns. In column four, AGGAN attends on the whole oranges while SPA-GAN has higher attention values around the boundaries and the top part of the oranges. Also note that the attention weights are generally greater than 0.5 as shown in the weight distributions in the third row. Thus, the generator will not produce agnostic outputs for the non-discriminative regions. }
\label{fig:Fig2}
\end{center}
\end{figure*}

In Fig.~\ref{fig:Fig2}, we show a few examples and compare our generated attention maps with the attention maps generated by the attention network in AGGAN \cite{mejjati2018unsupervised} on different datasets. Rows from top to bottom are the input images, attention maps computed in the discriminator of SPA-GAN, the distribution of weights in the attention maps of SPA-GAN, and the attention maps generated by the attention network in AGGAN \cite{mejjati2018unsupervised}, respectively. For instance, as shown in the fourth column of Fig.~\ref{fig:Fig2}, in the orange$\rightarrow$apple translation, the SPA-GAN attention map computed in the discriminator focuses on both the shape and texture of the generated and real apple images in order to correctly classify the inputs. It has higher values around the boundaries and on the top part of the oranges while AGGAN attends on the whole oranges. Moreover, as shown by the weight distributions in the third row of Fig.~\ref{fig:Fig2}, the SPA-GAN attention maps contain higher weights close to 1 for the most discriminative regions between the two domains and lower weights for other regions. It is important to note that the minimum value for the attention weights is around 0.5. These non-zero attention weights guarantee that the generator will not produce agnostic outputs for the non-discriminative regions. Transferring this knowledge to the generator improves the generator performance by focusing more on the discriminative areas and makes it more robust to shape changes between the two domains.

\begin{figure*}[!htbp]
\begin{center}
   \includegraphics[width=0.95\linewidth]{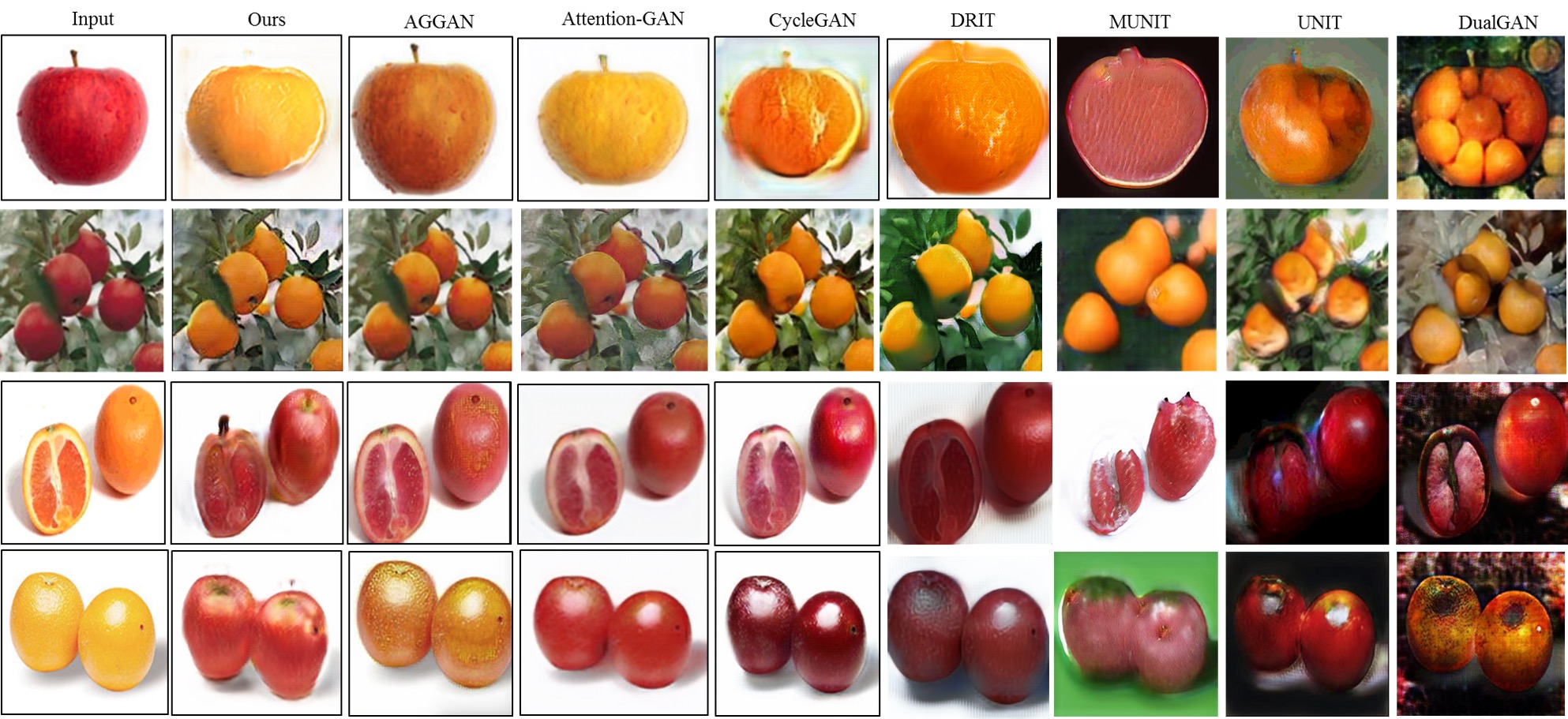}
                  \vspace{-0.3cm}
   \caption{Translation results generated by different approaches on the Apple$\leftrightarrow$Orange dataset.}
\label{fig:Fig3}
\end{center}
\end{figure*}

\begin{figure*}[!htb]
\begin{center}
   \includegraphics[width=0.95\linewidth]{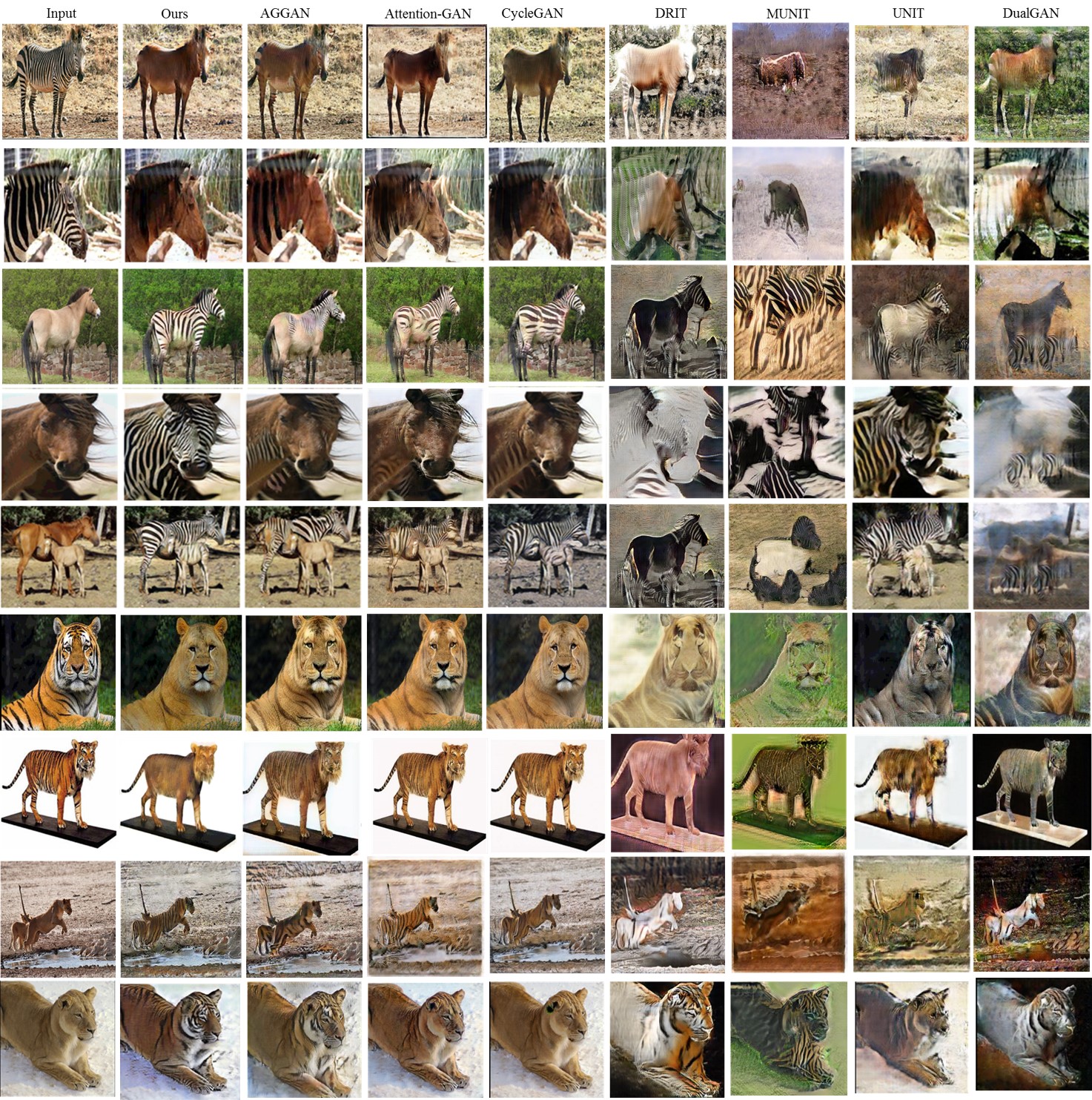}\\
   \setlength{\belowcaptionskip}{-20pt}
                     \vspace{-0.3cm}
   \caption{Image-to-image translation results generated by different approaches on the Zebra$\leftrightarrow$Horse and Tiger$\leftrightarrow$Lion datasets.}
\label{fig:Fig4}
\end{center}
\end{figure*}

Figs.~\ref{fig:Fig3} and~\ref{fig:Fig4} demonstrate some exemplar image-to-image translation results on benchmark datasets. The first column is the real input image and the generated images using SPA-GAN and other approaches are shown in the next columns. In all rows of Fig.~\ref{fig:Fig3}, DRIT, CycleGAN, Attention-GAN and AGGAN only changed the color of the objects and don't succeed in translating shape differences between apple and orange domains. As a comparison, SPA-GAN is more robust to shape changes and succeed on localizing parts of object and translating them to the target domain. It is clear that our approach generates more realistic images by changing both shape and texture of the input objects in the apple$\leftrightarrow$orange dataset. This validates the effectiveness of incorporating the attention to the generative network instead of applying the attention on the output of the transformation network \cite{mejjati2018unsupervised,chen2018attention}.

As shown in Fig.~\ref{fig:Fig4}, DualGAN, UNIT, MUNIT and DRIT altered the background of the input image. For example, the generated images by these methods in rows 3 and 4 have zebra patterns in the background. CycleGAN, Attention-GAN and AGGAN generate visually better results and preserve the input background. However, they miss certain parts of the object in the translation. For example, CycleGAN doesn't succeed to translate the head of zebra in rows 1 and 4 while AGGAN misses the body or the head of the animal in rows 1, 3 and 4. The generated objects by CycleGAN, Attention-GAN and AGGAN are mixed with parts from the target as well as the source domain. It can be seen in rows 3 and 5 that CycleGAN generates images with horizontal zebra patterns instead of vertical ones by SPA-GAN. In the tiger $\rightarrow$ lion translation, all other methods kept some tiger patterns after translation. SPA-GAN is also more successful in generating tiger pattern in lion $\rightarrow$ tiger translation (rows 8 and 9). Overall, SPA-GAN results are more realistic when compared with all other methods.  Please see the \textbf{supplementary material} for more visual examples.

\begin{table*}[!htb]
\centering
 \vspace{-0.3cm}
\caption{Kernel Inception Distance $\times$ 100 $\pm$ std. $\times$ 100 (lower is better) computed using only the target domain for various image-to-image translation methods on the Horse$\leftrightarrow$ Zebra, Apple $\leftrightarrow$ Orange and Tiger $\leftrightarrow$ Lion datasets.}.
 \vspace{-0.3cm}
\label{table:Table2}
 \vspace{-0.3cm}
\begin{tabular}{c|cccccc}
\hline
Method & apple $\rightarrow$ orange & orange $\rightarrow$ apple & zebra $\rightarrow$ horse & horse $\rightarrow$ zebra & lion $\rightarrow$ tiger & tiger $\rightarrow$ lion\\
\hline
DualGAN \cite{yi2017dualgan} & 14.68 $\pm$ 1.10 & 8.66 $\pm$ 0.94 & 9.82 $\pm$ 0.83 & 11.00 $\pm$ 0.68 & 11.5 $\pm$ 0.43 & 10.04 $\pm$ 0.76 \\
UNIT \cite{liu2017unsupervised} & 15.11 $\pm$ 1.41 & 7.26 $\pm$ 1.02 & 7.76 $\pm$ 0.80 & 6.35 $\pm$ 0.70 & 8.14 $\pm$ 0.25 & 8.17 $\pm$ 0.94\\
MUNIT \cite{huang2018multimodal} & 13.45 $\pm$ 1.67 & 6.79 $\pm$ 0.78 & 6.32 $\pm$ 0.90 & 4.76 $\pm$ 0.63 & 2.67 $\pm$ 0.63 & 8.10 $\pm$ 0.87 \\
DRIT \cite{lee2018diverse} & 9.65 $\pm$ 1.61 & 6.50 $\pm$ 1.16 & 5.67 $\pm$ 0.66 & 4.30  $\pm$ 0.57 & 2.39 $\pm$ 0.67 & 7.04 $\pm$ 0.73 \\
CycleGAN \cite{zhu2017unpaired} & 11.02 $\pm$ 0.60 & 5.94 $\pm$ 0.65 & 4.87 $\pm$ 0.52 & 3.94 $\pm$ 0.41 & 2.56 $\pm$ 0.13 & 5.32 $\pm$ 0.47\\

Attention-GAN \cite{chen2018attention} & 11.17 $\pm$ 0.92 & 5.41 $\pm$ 0.87 & 5.14 $\pm$ 0.68 & 4.67 $\pm$ 0.52 & 2.64 $\pm$ 0.24 & 6.28 $\pm$ 0.33 \\

AGGAN \cite{mejjati2018unsupervised} & 10.36 $\pm$ 0.86 & 4.54 $\pm$ 0.50 & 4.46 $\pm$ 0.40 & 4.12 $\pm$ 0.80 & 2.23 $\pm$ 0.21 & 5.83 $\pm$ 0.51 \\
SPA-GAN & \textbf{3.77 $\pm$ 0.32}  & \textbf{2.38 $\pm$ 0.33} & \textbf{2.19 $\pm$ 0.12} & \textbf{2.01 $\pm$ 0.13} & \textbf{1.17 $\pm$ 0.19} & \textbf{3.09 $\pm$ 0.19}\\
\hline
\end{tabular}
\end{table*}

\begin{table*}[!htb]
\begin{center}
\caption{Kernel Inception Distance $\times$ 100 $\pm$ std. $\times$ 100 (lower is better) computed using both the target and the source domains for various image-to-image translation methods on the Horse$\leftrightarrow$ Zebra, Apple $\leftrightarrow$ Orange and Tiger $\leftrightarrow$ Lion datasets.}.
\label{table:Table3}
\begin{tabular}{c|ccccccc}
\hline
Method & apple $\rightarrow$ orange & orange $\rightarrow$ apple & zebra $\rightarrow$ horse & horse $\rightarrow$ zebra & lion $\rightarrow$ tiger & tiger $\rightarrow$ lion\\
\hline
DualGAN \cite{yi2017dualgan} & 13.04 $\pm$ 0.72 & 12.42 $\pm$ 0.88 & 12.86 $\pm$ 0.50 & 10.38 $\pm$ 0.31 & 10.18 $\pm$ 0.15 & 10.44 $\pm$ 0.04\\
UNIT \cite{liu2017unsupervised} & 11.68 $\pm$ 0.43 & 11.76 $\pm$ 0.51 & 13.63 $\pm$ 0.34 & 11.22 $\pm$ 0.24 & 11.00 $\pm$ 0.09 & 10.23 $\pm$ 0.03\\
MUNIT \cite{huang2018multimodal} & 9.70 $\pm$ 1.22 & 10.61 $\pm$ 1.16 & 11.51 $\pm$ 1.27 & 8.31 $\pm$ 0.46  & 10.87 $\pm$ 0.91 & 10.61 $\pm$ 0.47 \\
DRIT \cite{lee2018diverse} & 6.37 $\pm$ 0.75 & 8.34 $\pm$ 1.22 & 9.65 $\pm$ 0.91 & 8.23 $\pm$ 0.08 & 9.56 $\pm$ 0.18 & 10.11 $\pm$ 0.59 \\
CycleGAN \cite{zhu2017unpaired} & 8.48 $\pm$ 0.53 & 9.82 $\pm$ 0.51 & 11.44 $\pm$ 0.38 & 10.25 $\pm$ 0.25 & 10.15 $\pm$ 0.08 & 10.97 $\pm$ 0.04\\

Attention-GAN \cite{chen2018attention} & 7.90 $\pm$ 0.25 & 8.05 $\pm$ 0.49 & 9.86 $\pm$ 0.32 & 8.28 $\pm$ 0.34 & 10.35 $\pm$ 0.58 & 10.56 $\pm$ 0.65\\

AGGAN \cite{mejjati2018unsupervised} & 6.44 $\pm$ 0.69 & \textbf{5.32 $\pm$ 0.48} & 8.87 $\pm$ 0.26 & \textbf{6.93 $\pm$ 0.27} & 8.56 $\pm$ 0.16 & 9.17 $\pm$ 0.07\\
SPA-GAN & \textbf{5.81 $\pm$ 0.51} & 7.95 $\pm$ 0.42 & \textbf{8.72 $\pm$ 0.24} & 7.89 $\pm$ 0.29  & \textbf{8.47 $\pm$ 0.07} & \textbf{8.63 $\pm$ 0.05}\\
\hline
\end{tabular}
\end{center}
\end{table*}

\begin{table*}[!htb]
\centering
\caption{Top-1 classification performance (higher is better) on images generated by various image-to-image translation methods on the Horse$\leftrightarrow$ Zebra, Apple $\leftrightarrow$ Orange and Tiger $\leftrightarrow$ Lion datasets.}.
\label{table:Table4}
\begin{tabular}{c|ccccccc}
\hline
Method & apple $\rightarrow$ orange & orange $\rightarrow$ apple & zebra $\rightarrow$ horse & horse $\rightarrow$ zebra & lion $\rightarrow$ tiger & tiger $\rightarrow$ lion \\
\hline
Real & 97.58  &  97.36 & 85.71  & 97.85 & 99.63 & 100 \\
\hline
DualGAN \cite{yi2017dualgan} &  78.57 & 64.91  &  41.42 & 83.33 & 66.53 & 39.05 \\
UNIT \cite{liu2017unsupervised} & 80.07  &  94.75 &  70.00 & 82.50  & 82.95 & 67.27 \\
MUNIT \cite{huang2018multimodal} & 67.80 & 85.70 & 55.27 & 82.50 & 79.60 & 52.75 \\
DRIT \cite{lee2018diverse} & 75.50 & 76.80 & 72.50 & 80.31 & 84.90 & 60.38 \\
CycleGAN \cite{zhu2017unpaired} & 71.80 & 72.93  & 75.00  & 83.33 & 73.48 & 48.10 \\

Attention-GAN \cite{chen2018attention} & 27.40 & 35.71 & 62.86 & 80.71 & 78.90 & 52.00 \\

AGGAN \cite{mejjati2018unsupervised} & 21.80 & 34.21  & 64.28 & 82.85 & 87.63 & 50.54 \\
SPA-GAN & \textbf{87.21} & \textbf{95.49} & \textbf{84.17} & \textbf{87.50} & \textbf{92.42} & \textbf{87.12} \\
\hline
\end{tabular}
\end{table*}

\subsection{Quantitative Comparison}
Mejjati et al. \cite{mejjati2018unsupervised} reported the mean KID value computed between generated samples using both source and target domains. We argue that calculating using both target and source domains is not a good practice especially for the datasets with no meaningful background such as Apple$\leftrightarrow$Orange. Therefore, we report mean KID values computed only on the target domain (Table~\ref{table:Table2}) and on both source and target domains (Table~\ref{table:Table3}) to better evaluate the performance of our proposed approach and state-of-the-arts.

Our approach achieved the lowest target only KID scores in all translation tasks, showing its effectiveness in generating more realistic images. It is interesting to see that SPA-GAN does not always achieve the smallest values when KIDs are computed with both the source and target. For example, in the column 2 of Table~\ref{table:Table3} (Orange $\rightarrow$ Apple), AGGAN has the smallest KID value of 5.32, which is averaged between the generated apples and real apples (target), and the generated apples and real oranges (source). As a comparison, SPA-GAN has the smallest KID value in column 2 of Table~\ref{table:Table2}, computed only based on the real apples. This clearly shows that the apples generated by AGGAN still maintain a higher level of feature similarity to real oranges when compared to SPA-GAN. That is, SPA-GAN results are more realistic. Results from Tables~\ref{table:Table2} and ~\ref{table:Table3} clearly demonstrate the effectiveness of SPA-GAN.
We also report the top-1 classification performance on the real images as well as the generated images by each method in Table~\ref{table:Table4}. If the generated image is real enough, the classifier will predict it as a target sample from the target domain. The images produced by SPA-GAN network clearly outperforms all competing methods in terms of classification accuracy.

\subsection{User Study Evaluation}
We further evaluated our method on apple $\leftrightarrow$ orange, horse $\leftrightarrow$ zebra and lion $\leftrightarrow$ tiger with a human perceptual study. In each task, we randomly selected 100 images from the test set, including 50 for forward mapping and 50 for the inverse. 10 participants are asked to select the most realistic image from images generated by CycleGAN, AGGAN and SPA-GAN (in random order).

Table~\ref{table:Table5} shows the participant votes for each task. Specifically, Table V provides the number of points received by CycleGAN, AGGAN and SPA-GAN, respectively, for three tasks. There are 100 images, and the winning method for each image receives one point. In some cases, two methods can get an equal number of votes from the 10 participants. In these scenarios where there is a tie, both tied winners receive 0.5 points. Clearly, SPA-GAN performs the best in the user study, which is consistent with the reported KID and classification accuracy.

\begin{table}
\centering
\caption{User study on Apple $\leftrightarrow$ Orange, Horse $\leftrightarrow$ Zebra and Lion $\leftrightarrow$ Tiger.}
\label{table:Table5}
\begin{tabular}{c|ccc}
\hline
Method & apple $\leftrightarrow$ orange & horse $\leftrightarrow$ zebra & lion $\leftrightarrow$ tiger\\
\hline
CycleGAN \cite{zhu2017unpaired} & 12  & 13.5 & 13  \\
AGGAN \cite{mejjati2018unsupervised} & 15  & 18 & 16.5 \\
SPA-GAN & \textbf{73} & \textbf{68.5} & \textbf{70.5} \\
\hline
\end{tabular}
\end{table}

\subsection{Other Image-to-Image Translation Applications}
\begin{figure}[htbp]
\centering
   \includegraphics[width=0.6\linewidth]{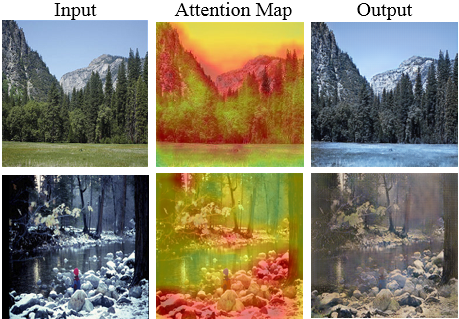}\\
   \caption{Translation results on the Winter$\leftrightarrow$Summer dataset.}
\label{fig:Fig5}
\end{figure}

\begin{figure}[htbp]
\centering
   \includegraphics[width=0.6\linewidth]{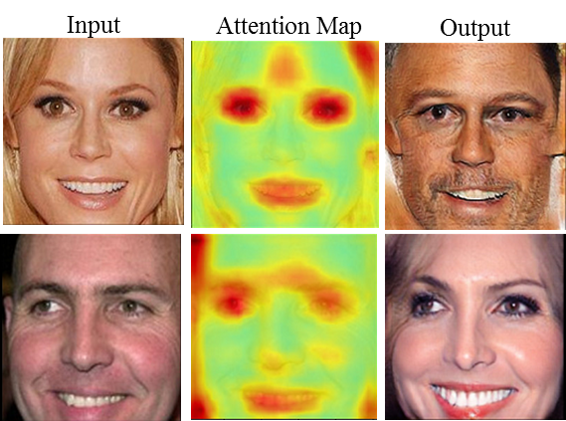}\\
   \caption{Translation results on gender conversion.}
\label{fig:Fig6}
\end{figure}

Finally, we evaluated SPA-GAN on image-to-image translation datasets that require to translate the whole image. In Fig.~\ref{fig:Fig5}, we show the results on the Winter $\leftrightarrow$ Summer dataset \cite{zhu2017unpaired}, where the second column shows the attention maps associated with the holistic translation of input images with no specific type of objects. Clearly, the discriminator focuses on the areas such as ground and trees that have different colors during the winter and summer seasons. In Fig.~\ref{fig:Fig6}, we show the gender conversion results on the Facescrub \cite{ng2014data} dataset. The attention maps in the second column show higher activations around different areas of the face such as eyes, nose and lips. These spatial attention maps clearly demonstrate the effectiveness of SPA-GAN in a variety of holistic image-to-image translation tasks. Please see the \textbf{supplementary material} for more visual examples on the Facescrub and GTA $\leftrightarrow$ Cityscapes datasets.

\section{Conclusion}
In this paper, we proposed SPA-GAN for image-to-image translation in unsupervised settings. In SPA-GAN, we compute the spatial attention maps in the discriminator and transfer the knowledge to the generator so that it can explicitly attend on discriminative regions between two domains and thus improve the quality of generated images. SPA-GAN is a lightweight model and achieved superior performance, both qualitative and quantitative, over current state-of-the-arts.

\ifCLASSOPTIONcaptionsoff
  \newpage
\fi

\bibliographystyle{IEEEtran}
\bibliography{ref}

\clearpage
\begin{figure*}
\textbf{Appendix}\par\medskip
\textbf{A. Additional experimental results}\par\medskip
\begin{center}
   \includegraphics[width=1\linewidth]{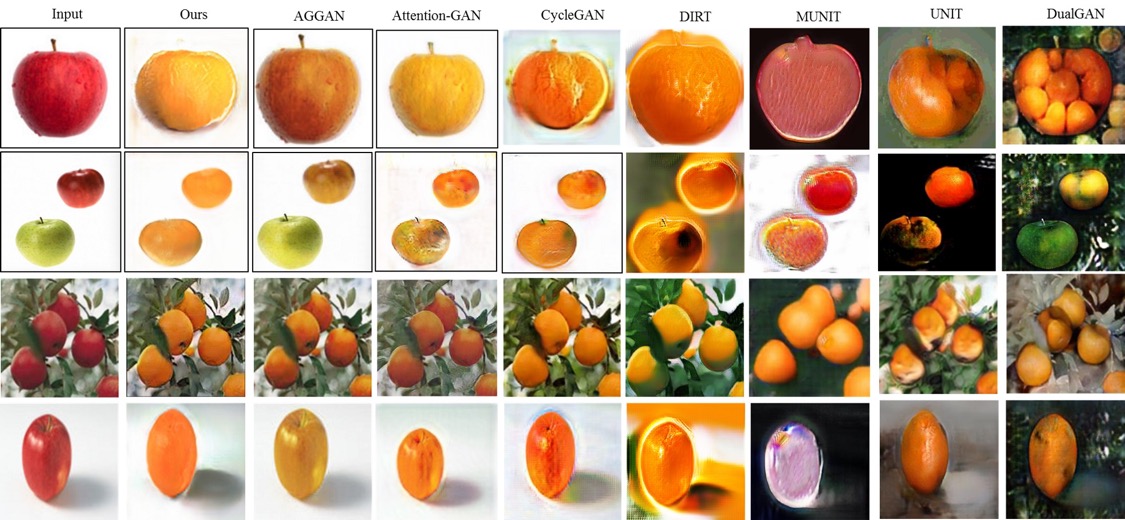}\\
   \caption{Apple $\rightarrow$ Orange translation results. DualGAN, UNIT and MUNIT altered the background of the input image and
do not succeed in translation. DRIT, CycleGAN, Attention-GAN and AGGAN only changed the color of the objects and
do not succeed in translating shape differences between apple and orange domains. As a comparison, SPA-GAN is more
robust to shape changes.}
\label{fig:Fig7}
\end{center}
\end{figure*}

\begin{figure*}
\begin{center}
   \includegraphics[width=1\linewidth]{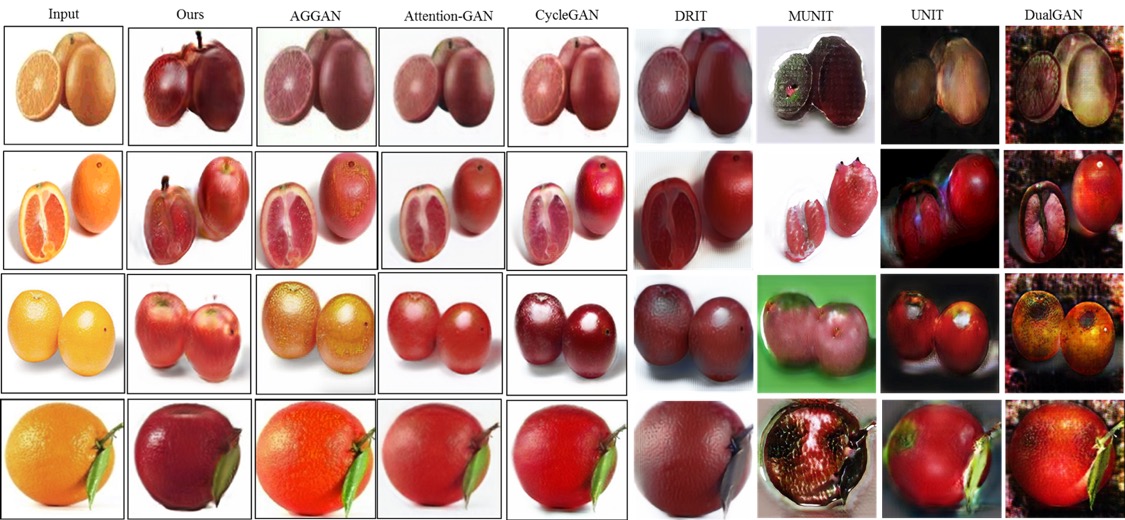}\\
   \caption{Orange $\rightarrow$ Apple translation results. DualGAN, UNIT and MUNIT altered the background of the input image and
do not succeed in translation. DRIT, CycleGAN, Attention-GAN and AGGAN only changed the color of the objects and
do not succeed in translating shape differences between apple and orange domains. As a comparison, SPA-GAN is more
robust to shape changes..}
\label{fig:Fig8}
\end{center}
\end{figure*}

\begin{figure*}
\begin{center}
   \includegraphics[width=1\linewidth]{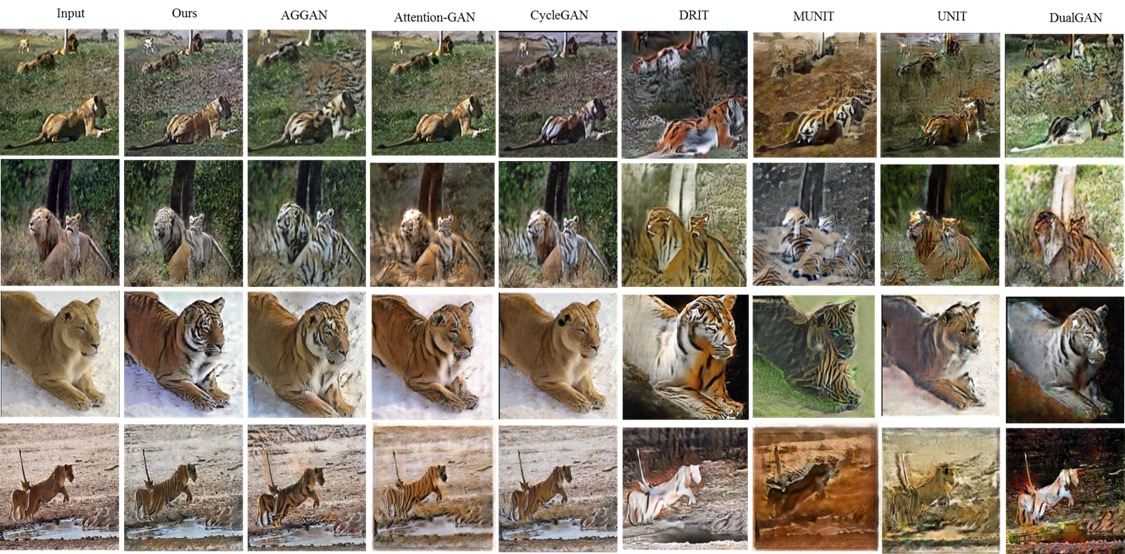}\\
   \caption{Lion $\rightarrow$ Tiger translation results. AGGAN, Attention-GAN and CycleGAN altered the background of the input
images (the generated images by AGGAN in row 1 and 2 have tiger patterns in the background). Clearly, SPA-GAN is
more successful in generating tiger pattern in row 3 and 4 compared to all other methods. }
\label{fig:Fig9}
\end{center}
\end{figure*}

\begin{figure*}
\begin{center}
   \includegraphics[width=1\linewidth]{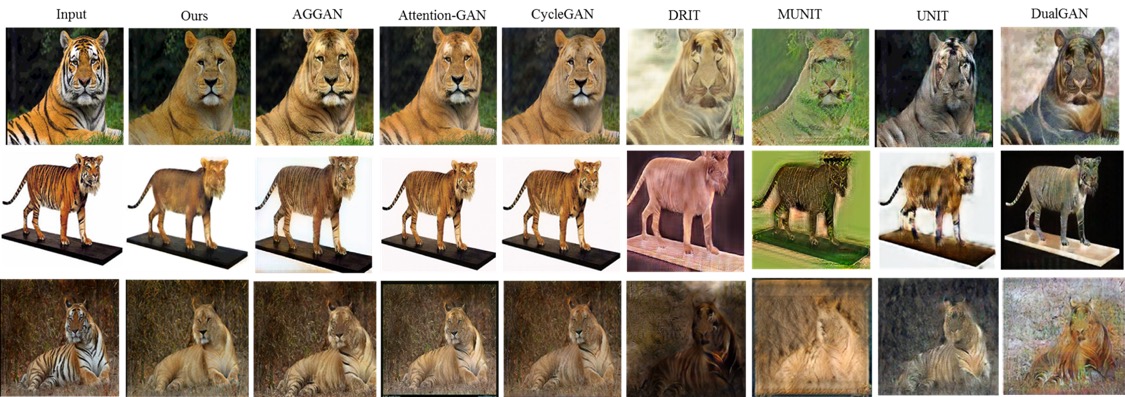}\\
   \caption{Tiger $\rightarrow$ Lion translation results. All other methods kept some tiger patterns after translation. DRIT, MUNIT, UNIT
and DualGAN altered the background of the input images in row 2. Clearly, SPA-GAN results are more realistic compared
to other approaches.}
\label{fig:Fig10}
\end{center}
\end{figure*}

\begin{figure*}
\begin{center}
   \includegraphics[width=1\linewidth]{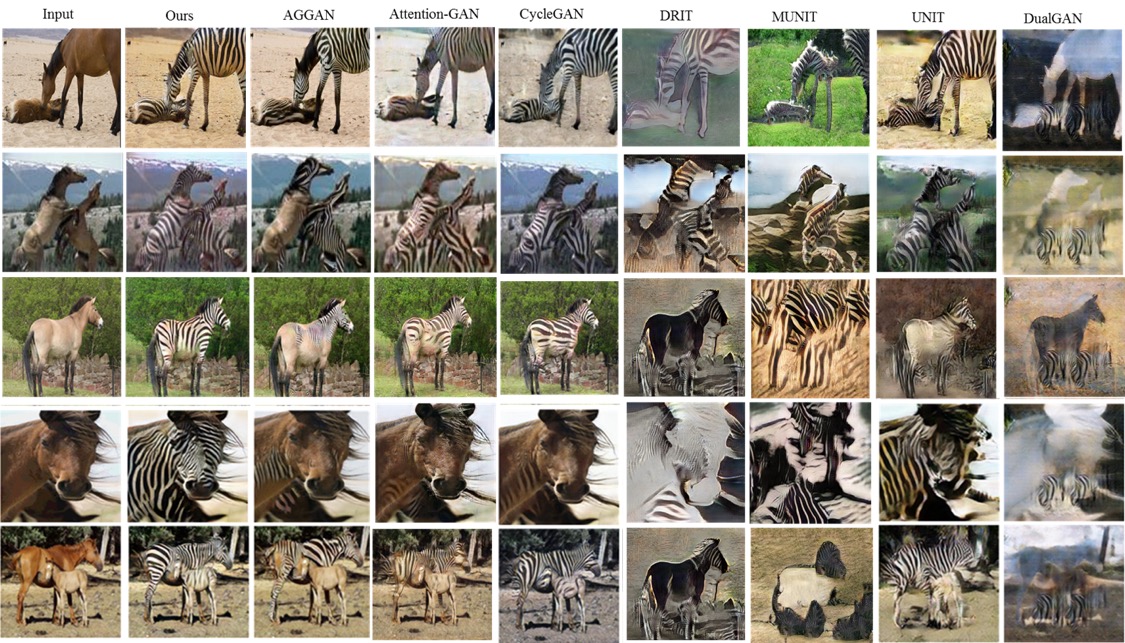}\\
   \caption{Horse $\rightarrow$ Zebra translation results. CycleGAN, Attention-GAN and AGGAN miss certain parts of the object in the
translation in row 1, 2, 4 and 5. Attention-GAN and CycleGAN generate images with unnatural skin pattern (horizontal
patterns) in row 2, 3 and 5. The generated objects by CycleGAN, Attention-GAN and AGGAN are mixed with parts from
the target as well as the source domain. }
\label{fig:Fig11}
\end{center}
\end{figure*}

\begin{figure*}
\begin{center}
   \includegraphics[width=1\linewidth]{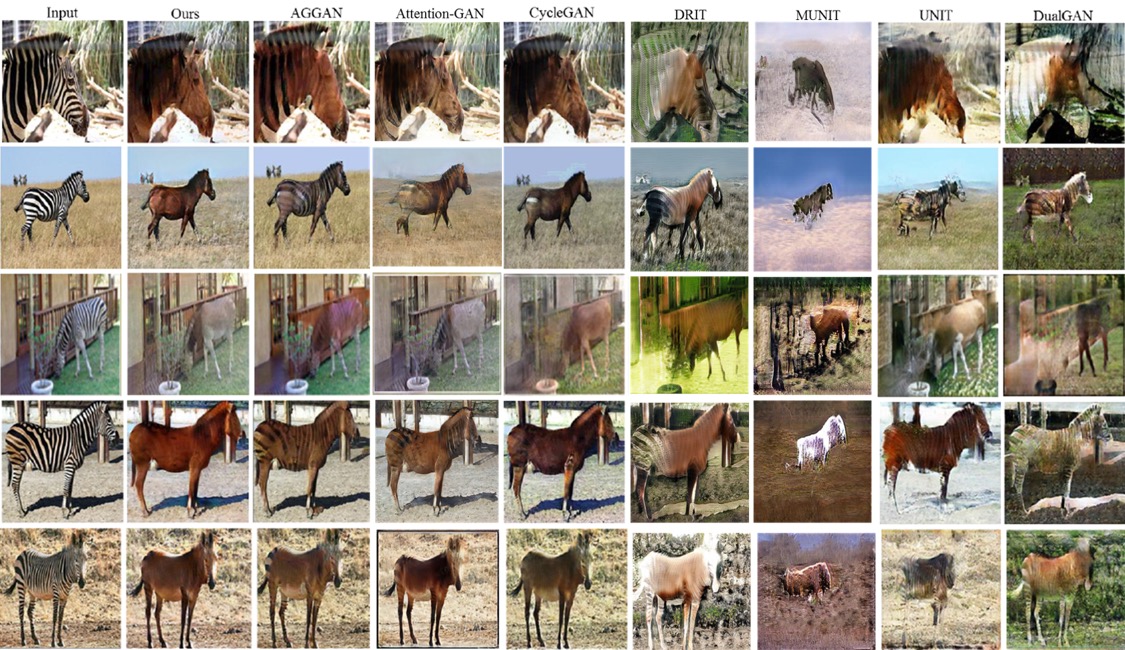}\\
   \caption{Zebra $\rightarrow$ Horse translation results. DualGAN, UNIT, MUNIT and DRIT altered the background of the input image
and do not succeed in translation. CycleGAN, Attention-GAN and AGGAN miss certain parts of the object in the translation.
They also kept some zebra patterns after translation in row 1, 2 and 4. In row 3, AGGAN, Attention-GAN and CycleGAN
fail to detect the zebra as foreground, and so change the background image content (fence) while SPA-GAN detects the
zebra and translates it to the target domain.}
\label{fig:Fig12}
\end{center}
\end{figure*}

\begin{figure*}
\begin{center}
   \includegraphics[width=0.8\linewidth]{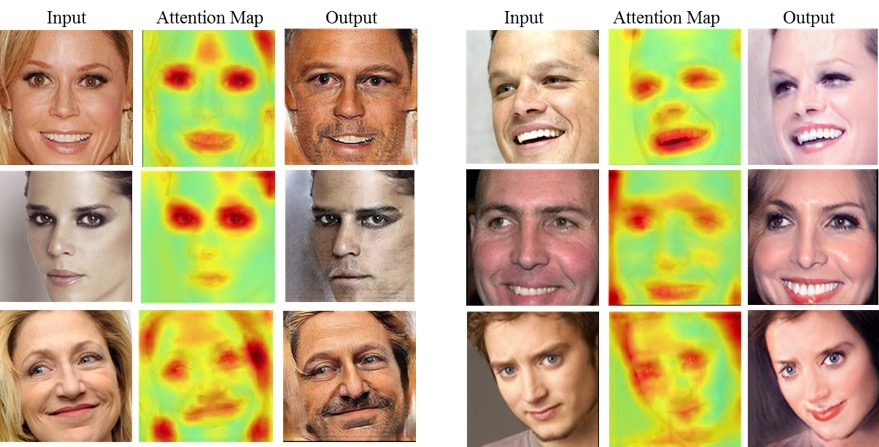}\\
   \caption{Translation results on the gender conversion (Facescrub dataset) requiring holistic translation for the input image with no specific type of object. In each group from left to right are the input images, the attention maps, and the translated images. The second and fifth columns show the attention maps with higher activation level around different areas of the face such as eyes, nose and lips that the discriminator attends to classify the input image.}
\label{fig:Fig13}
\end{center}
\end{figure*}

\begin{figure*}
\begin{center}
   \includegraphics[width=1\linewidth]{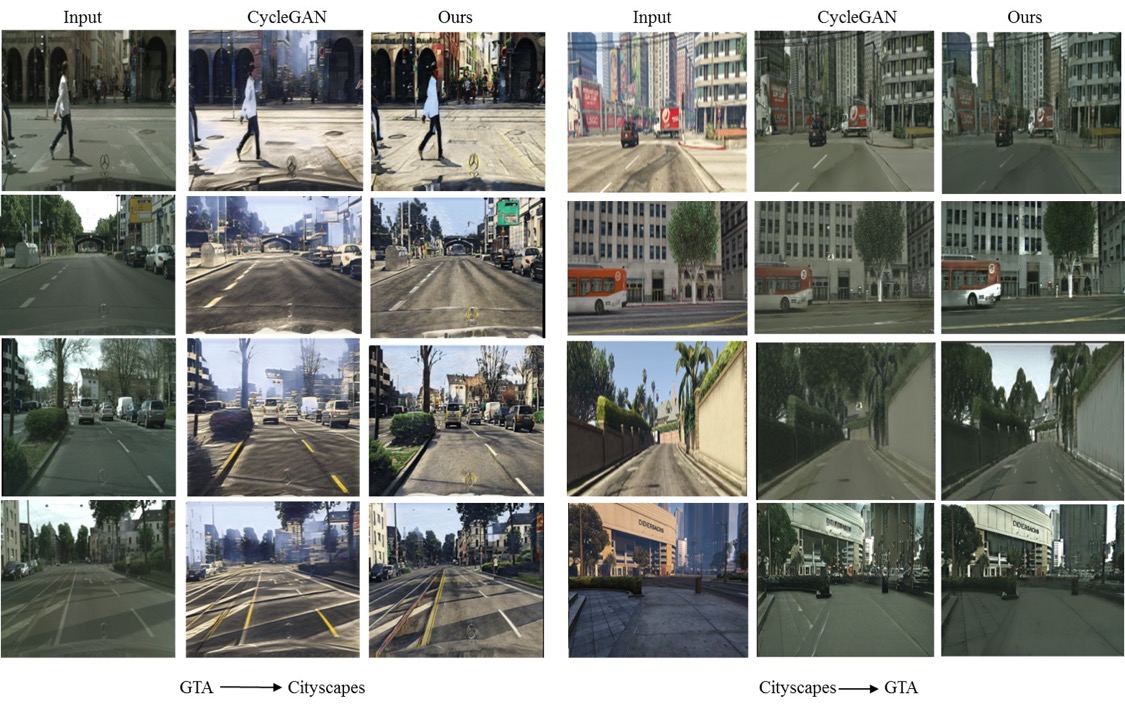}\\
   \caption{Translation results on GTA  $\leftrightarrow$ Cityscapes requiring holistic translation for the input image with no specific type of object.}
\label{fig:Fig14}
\end{center}
\end{figure*}

\end{document}